\documentclass[letterpaper, 10 pt, conference]{ieeeconf}
\usepackage[utf8]{inputenc}
\usepackage[english]{babel}
\usepackage{csquotes}
\usepackage[
backend=bibtex,
style=numeric,
natbib=true, 
isbn=false,
doi=false
]{biblatex}
\usepackage{graphicx,subcaption} %
\IEEEoverridecommandlockouts
\usepackage{xcolor}
\usepackage{acronym}
\usepackage{lipsum}

\title{\LARGE \bf
Robo-PlaNet: Learning to Poke in a Day
}

\author{Maxime Chevalier-Boisvert*$^{1}$,  Guillaume Alain*$^{1}$, Florian Golemo$^{1,2}$, and Derek Nowrouzezahrai$^{1,3}$%
\thanks{* Equal contribution, contact: maxime.chevalier-boisvert@mila.quebec}%
\thanks{$^{1}$Mila - Quebec AI Institute}%
\thanks{$^{2}$ElementAI}%
\thanks{$^{3}$McGill}%
}

\acrodef{MPC}{Model Predictive Control}
\acrodef{DoF}{degrees of freedom}
\acrodef{VAE}{Variational Autoencoder}
\acrodef{CEM}{the Cross-Entropy Method}
\acrodef{RL}{Reinforcement Learning}
\acrodef{RNN}{recurrent neural network}

\begin{document}

\maketitle

\begin{abstract}
Recently, the Deep Planning Network (PlaNet) approach was introduced as a model-based reinforcement learning method that learns environment dynamics directly from pixel observations. This architecture is useful for learning tasks in which either the agent does not have access to meaningful states (like position/velocity of robotic joints) or where the observed states significantly deviate from the physical state of the agent (which is commonly the case in low-cost robots in the form of backlash or noisy joint readings). PlaNet, by design, interleaves phases of training the dynamics model with phases of collecting more data on the target environment, leading to long training times. In this work, we introduce Robo-PlaNet, an asynchronous version of PlaNet. This algorithm consistently reaches higher performance in the same amount of time, which we demonstrate in both a simulated and a real robotic experiment.
\end{abstract}

\section{Introduction}

Teaching a robot a new trick can prove challenging. Currently, many methods rely on transfer from simulation to physical platforms (i.e. "sim2real transfer"). However, no simulation is perfect and despite recent advances in methods to make the transfer more robust (e.g. \cite{mehta2019active,golemo2018sim}), there remains a performance gap between robots trained on simulated and real data. This can, for example, be due unmodelled or hard-to-model effects like backlash, or due to interactions with dynamic objects. For these tasks, in order to learn accurate dynamics, real robot data is necessary. 

Recently \citet{planet} proposed an approach (Deep Planning Networks or ``PlaNet'') to learn an environment's dynamics directly on the target task. Their proposed method learns a transition model of the robot's environment during training and uses this for \ac{MPC}. However, the authors have only demonstrated their work in simulation. We implemented their work on a real robot and found it to be impractical. Most physical robots need supervision during training. The original PlaNet algorithm interleaves episodes of robot rollouts with long periods of network training during which the robot remains idle and the robot's overseer has to wait patiently.

In this work, we introduce Robo-PlaNet, an asynchronous version of PlaNet, which trains faster by minimizing robot downtime. We evaluate this on a robotic reaching task, which we can learn to solve in a fraction of the time taken by vanilla PlaNet.

There are currently few methods that are able to learn a robotic policy directly on live robots. Existing works like PILCO~\cite{deisenroth2011pilco} either work only in low-dimensional settings (i.e. from robotic joint sensors, but not from images) or only work on a very limited set of tasks where the states between start and goal can be linearly interpolated \cite{singh2019}. The original PlaNet algorithm does not suffer from either of these shortcomings since it learns a general visual predictive model of environment dynamics. To illustrate the practicality of our approach, we implemented our reaching task on a low cost Poppy Ergo Jr. robot~\cite{lapeyre2014poppy}, which is known to have complex dynamics due to backlash~\cite{golemo2018sim}.

\begin{figure}[bt]
\begin{center}
\includegraphics[width=.45\textwidth]{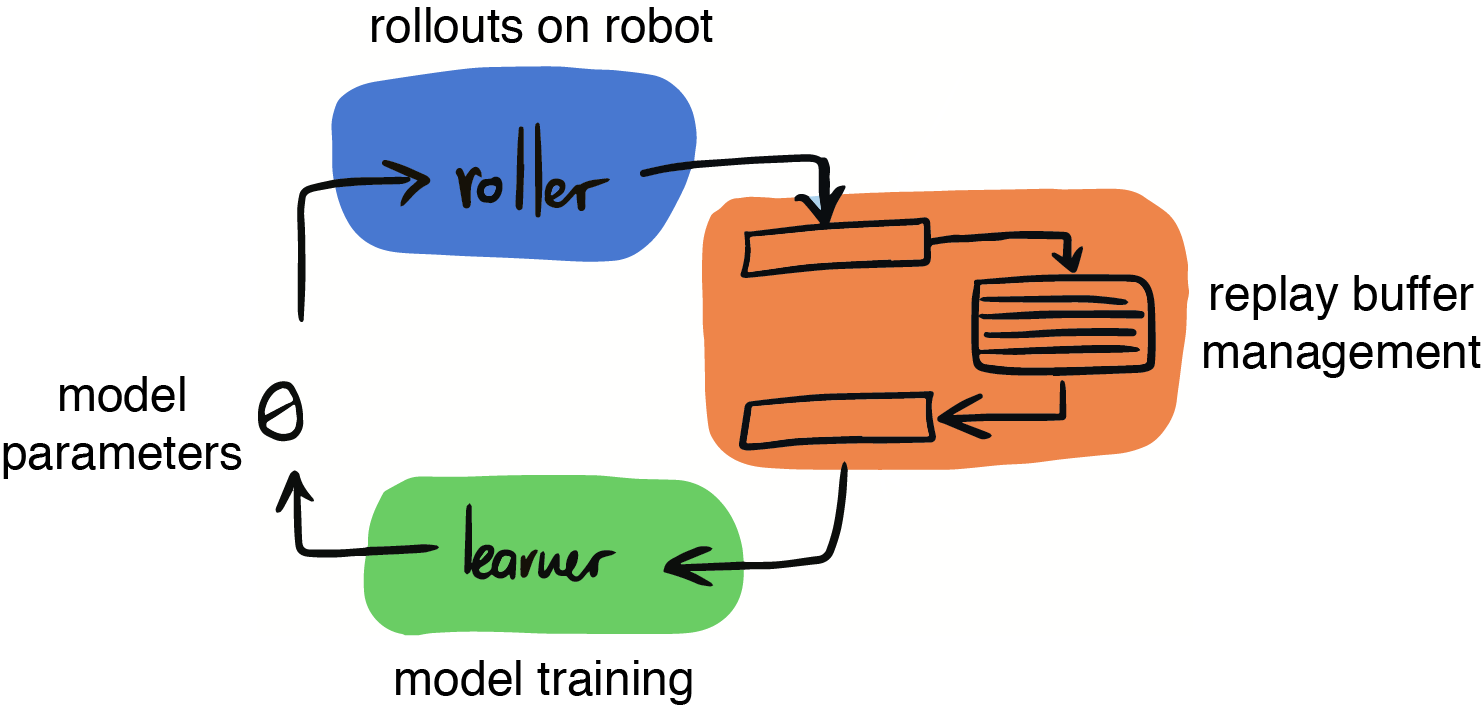}
\caption{\textbf{Overview Robo-PlaNet}. Illustration of the change over the original PlaNet architecture~\cite{planet}. We parallelize learning and gathering robot rollouts. The rest of the training process is identical to PlaNet. With this simple modification, we achieve much faster training and minimize operator time.}
\vspace{-.5em}
\label{fig:cycle}
\end{center}
\end{figure}

Our pipeline uses an OpenAI Gym-like robotic environment~\cite{brockman2016openai}, in which the environment calculates a dense reward for a given state and action. We use 2 off-the-shelf RGB webcams instead of a single RGB-D camera in our reaching task to allow for better visibility of both end-effector and target object. Both input images are encoded separately and fused late in our network. In our system, there are 3 different processes running at all times: one continuously gathering data using the latest model on the robot, one for training new models from collected experience, and one for managing the replay buffer (see Fig.~\ref{fig:cycle}).

\begin{figure*}[tb!]
  \begin{subfigure}{0.5\textwidth}
    \centering
    \includegraphics[width=.95\textwidth]{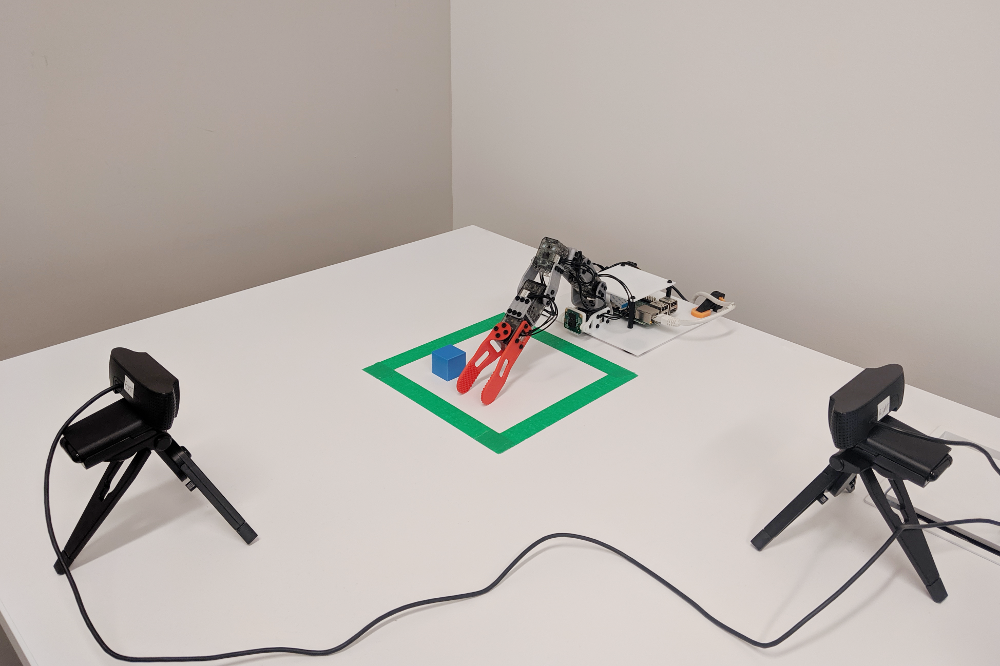}
    \caption{Setup for the reaching task}
    \label{fig:setup}
  \end{subfigure}%
  \begin{subfigure}{0.5\textwidth}
    \begin{subfigure}{\textwidth}
      \centering
      \includegraphics[width=.63\textwidth]{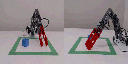}
    \end{subfigure}
    \begin{subfigure}{\textwidth}
      \centering
      \includegraphics[width=.63\textwidth]{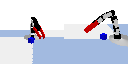}
    \end{subfigure}
    \caption{Sample observations}
    \label{fig:obs}
  \end{subfigure}
  \caption{\textbf{Experimental Setup.} In this task, the robot has to touch the object that is randomly placed in front of it via torque control. Left: two-camera setup showing cube placement area, Right: real and simulated robot observations side-by-side. Note that we are evaluating our method in these two environments separately - we are not transferring the policy from the simulation to the real robot.}
  \label{fig:setup-obs}
\end{figure*}

Our experimental setup consists of the 6 \ac{DoF} Poppy Ergo Jr. robot arm with a gripper end effector, a 3D printed cube as target, and two fixed external cameras. The goal of the robot is to learn how to reach and touch the target cube which is randomly moved to different positions in front of the robot (see Fig.~\ref{fig:setup-obs}). The experiments were carried out both in simulation and on the physical platform. In both cases, our results indicate a noticeable reduction in training time until convergence.

The main contribution of this work is to extend PlaNet to asynchronous/parallel operation to improve training time.

\section{Background}

PlaNet consists of 3 networks: (a) a \ac{VAE} that encodes an image observation into a latent state; (b) a reward estimator that learns the reward that is associated with each latent state; and (c) a \ac{RNN} which learns to predict the next latent state given the previous one and an action. 
Together, these components can be used to encode a single observation, roll out multiple alternative trajectories in latent space without interacting with the environment, and then estimate their cumulative reward several steps into the future. 

The authors combine their method with \ac{MPC} to find optimal actions over a finite planning horizon. This planning is repeated at every timestep. 

PlaNet is initialized with random trajectories to create an initial model of the environment dynamics. After that, phases of fitting the three networks to the observed data and phases of gathering new data by following the current policy are alternated. The data is stored in and randomly sampled from a replay buffer, which makes this method off-policy. As mentioned above, if executed on a physical robot, this causes the robot to pause while the networks are being trained.

At the same time as this work, another conceptually similar work was released: \citet{zhang2019asynchronous}. The authors show how asynchronous methods can speed up training of a variety of different contemporary \ac{RL} methods. The idea of parallelizing deep \ac{RL} methods and having policies train while new data is being collected is not novel in itself (see e.g. \cite{kalashnikov2018qt,nair2015massively}). In this work, however, our contribution does not lie in providing the fastest possible training algorithm, but in using PlaNet as a starting point and improving the algorithm for real-world application.

\section{Method}

The original PlaNet approach alternates between data collection and model training. When new trajectories are collected in the environment, they are stored in a replay buffer and later used for training the models. During model training, no interaction with the environment takes place.

Creating this work, our hypothesis was that the alternation between collection and training phases was chosen for implementation convenience and not inherently necessary. With our approach, environment interaction does not take place with models that had access to the most recent trajectories. However, \citet{luo2018algorithmic} argued that this acts beneficially as a policy regularizer. 

To build Robo-PlaNet, we factored PlaNet\footnote{We used the PyTorch implementation from \url{https://github.com/Kaixhin/PlaNet}.} into three separate processes running independently (see Fig.~\ref{fig:cycle}). Theses processes are spawned by the PyTorch multiprocessing library and communicate via queues provided by the same. This approach is not conceptually dependent on PyTorch and would work similarly with TensorFlow or other libraries. We distinguish the following processes:

\begin{itemize}
    \item \emph{Roller Process (GPU)}: This process interacts with the environment. It is spawned with access to a queue for incoming model parameters. After each episode, as soon as there is one or more items in the queue, the process extracts the latest element, discards all other elements in the queue, and uses it to update the model. The rollouts generated by this process are sent to the queue of the Replay Buffer Process.
    \item \emph{Replay Buffer Process (CPU)}: This process manages all the stored trajectories. It reads the trajectories from a queue, stores them into memory, samples minibatches uniformly, and writes the samples to the queue for the Learner Process.
    \item \emph{Learner Process (GPU)}: This process trains the \ac{VAE}, the reward estimator, and the \ac{RNN}-based state transition model. It reads minibatches from the queue, updates the model parameters according to \cite{planet}, and finally pushes the parameters of the trained model into the queue for the \emph{Roller Process}.
\end{itemize}

The original PlaNet implementation uses a \ac{VAE} to encode the pixels from the single camera into a latent state distribution. Since our task requires 2 cameras, we encode them separately, and concatenate the latents before sampling from them. We also employ 2 separate decoders that restore the 2 camera images based on the same latent code. We believe that this late fusion approach can easily be extended to 3 or more cameras by just including more encoders/decoders and concatenating/splitting the latents as we do in our approach.

We deliberately moved the Roller and Learner processes onto the GPU since one trains the neural network models and the other needs to run the model in real-time in order to carry out actions on the physical robot.

The format of the replay buffer is similar to the original PlaNet implementation with the exception that the images from both cameras are concatenated horizontally (i.e. $128\times64$ pixels as opposed to $64\times64$).

\section{Experiments}

\begin{figure*}[tb!]
  \begin{subfigure}{0.5\textwidth}
    \centering
    \includegraphics[width=.95\textwidth]{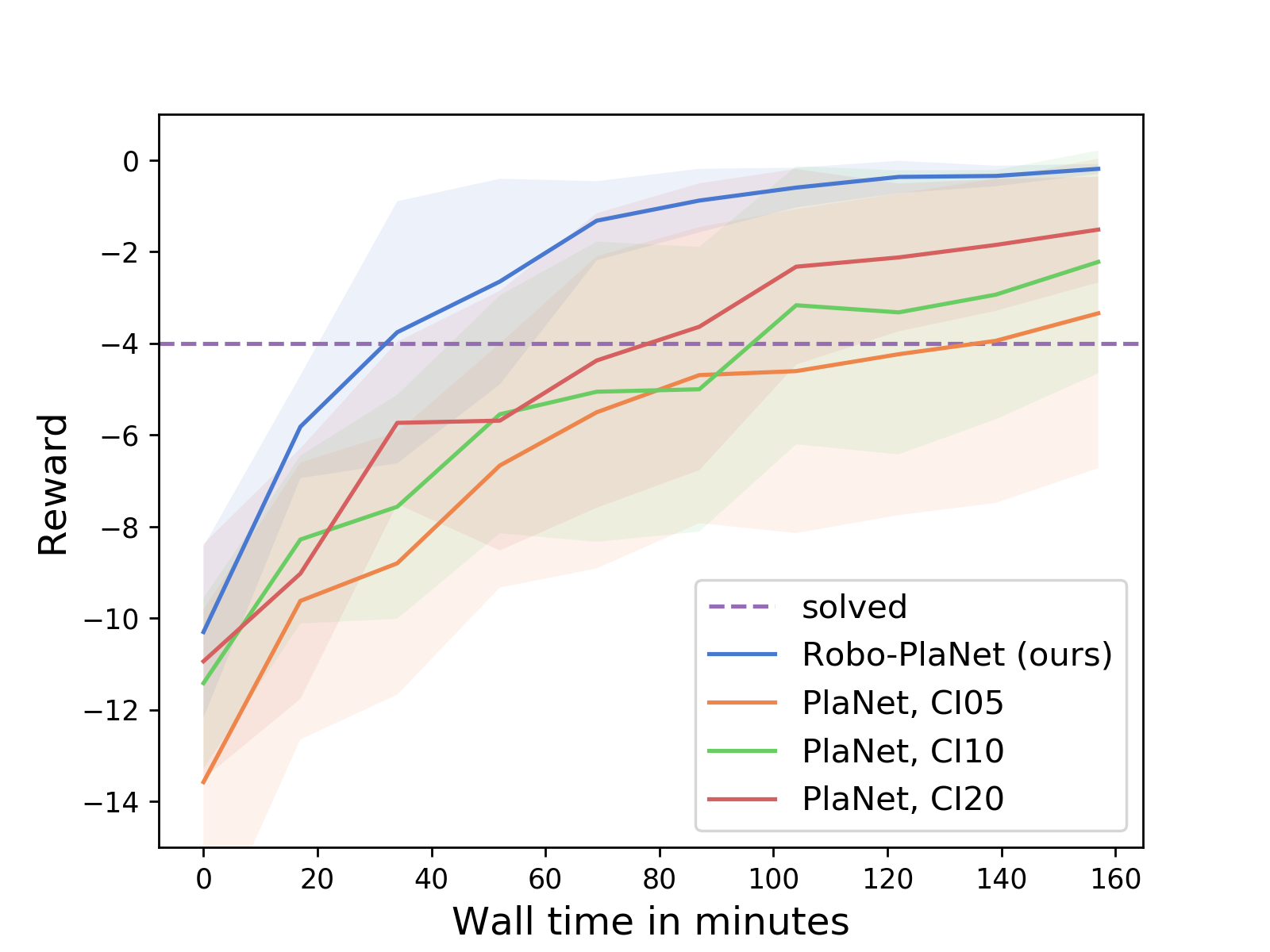}
    \caption{Reward over time, in simulation}
    \label{fig:results-sim}
  \end{subfigure}%
  \begin{subfigure}{0.5\textwidth}
    \centering
    \includegraphics[width=.95\textwidth]{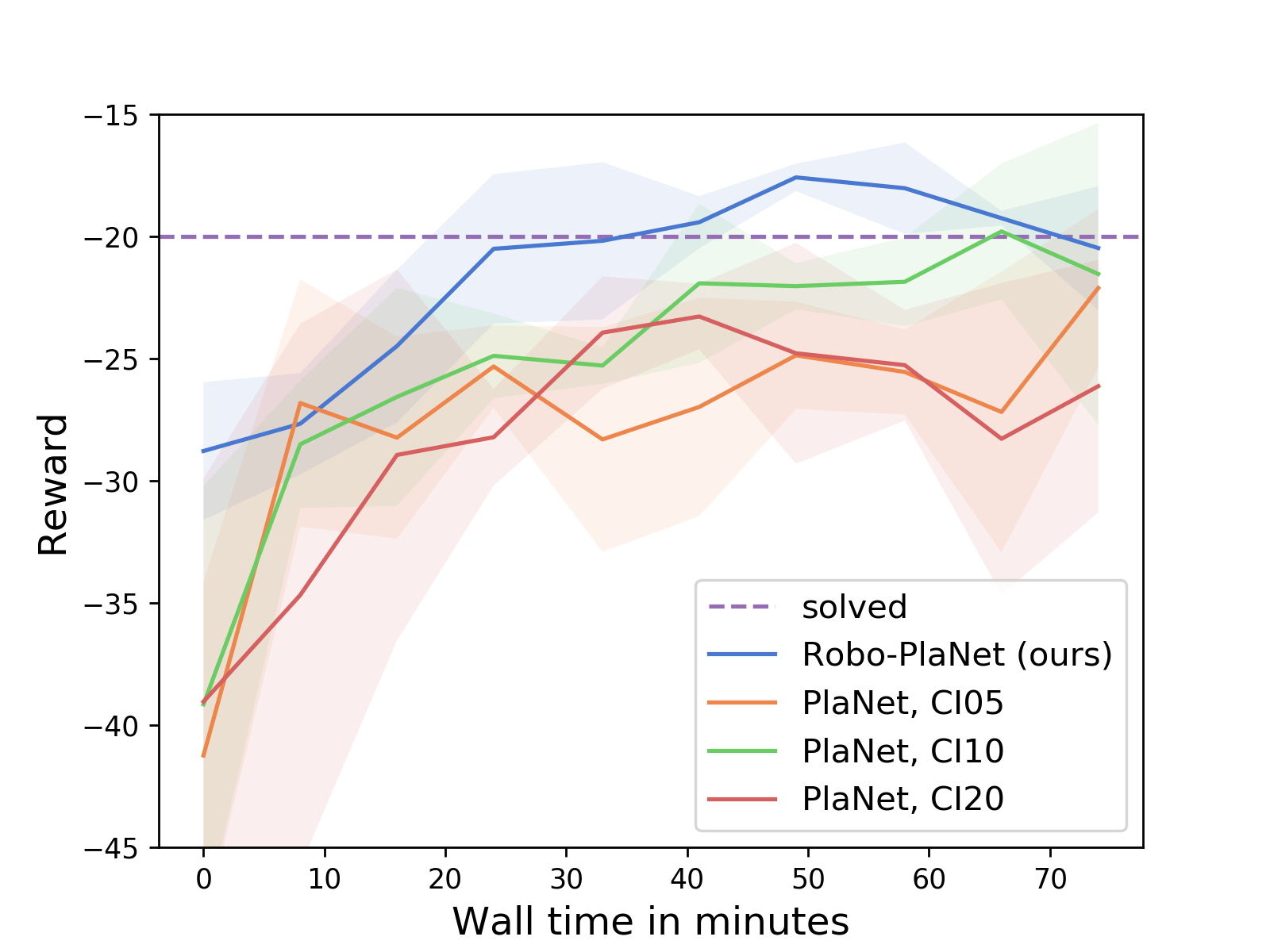}
    \caption{Reward over time, on real robot}
    \label{fig:results-real}
  \end{subfigure}%
  \caption{\textbf{Results.} Performance during training for the unmodified PlaNet implementation (under different hyperparameters. collect interval 5, 10, 20 as indicated by ``CI05'', ``CI10'', ``CI20'', respectively) and our approach (``Robo-PlaNet''). Reward was smoothed via data binning. In both simulated and real environments, our algorithm reaches the point of solving the environment, i.e. poking the cube, sooner.}
  \label{fig:results}
\end{figure*}

The following section outlines the experiments that were carried out in simulation and on the real robot. 

\subsection{Experimental Setup}

For evaluating our algorithm, we are using the 6-\ac{DoF} Poppy Ergo Jr robot\footnote{During our experiments we kept the 4th joint fixed and the gripper's movement very limited, thus making this effectively a 4-\ac{DoF} task.} in a reaching task. In this task, the robot has to reach a cube that is randomly placed in front of it, as fast as possible. Each episode runs for 100 steps (in simulation, we also terminate an episode when the cube is touched). Upon resetting, the arm moves to a resting position plus uniform noise ($0 \pm 18^\circ$). The arm is controlled via torque control, i.e. similarly to the OpenAI Gym environment ``Reacher-v2''. The observations are 2 images measuring $64\times64$ pixels, horizontally concatenated from both external cameras. The cameras are arranged directly in front of the robot and 90 degrees to the right, as shown in Fig.~\ref{fig:setup}. The reward is the inverse distance of the end effector to the cube. We have chosen the 2 camera setup because with a single camera it was possible for the robot to obstruct view of the cube. The simulation\footnote{Available at \url{https://github.com/fgolemo/gym-ergojr/}} is implemented in PyBullet\cite{coumans2019}. Non-stationary behaviors that are present on the real robot like backlash and heat-dependent acceleration were not modeled. While in simulation the reward is calculated directly as the distance between object centers of end effector and cube, on the real robot it is the inverse of the sum of pixel distances in each camera frame. After each episode on the physical system, the operator moves the cube to a random position that is indicated on the terminal. 

In traditional RL settings, it is common to train for a number of frames. However, in our case we would like to compare by wall clock time and correspondingly, we ran all tasks for a fixed number of minutes instead of frames. The simulated tasks ran for 180 minutes over 7 seeds and the real robot experiments for 90 minutes over 3 random seeds per policy type. The goal of these experiments was to compare performance (as measured by average episode reward) after a given period of training and interaction with the environment. Our hypothesis was that the asynchronous environment would be able to collect rollout data faster and thus allow for better model performance over the same period of time. 

The original PlaNet implementation contains a ``collect interval'' hyperparameter that specifies how many episodes the model is trained for before interacting with the environment again. In our Robo-PlaNet framework, this does not matter since the model training and the rollout collection are independent but in the unmodified PlaNet algorithm, this is an important hyperparameter. Therefore, in order to provide fair comparison, we compared our model with 3 different values for the collect interval (5, 10, and 20).

All experiments were carried out on a computer with 2 Nvidia GTX 1080 GPUs. One GPU was used for planning and experience collection, while the other one was used to train the models.

\subsection{Results}

The results are displayed in Fig.~\ref{fig:results}. The data was binned  with 10 bins for plotting since it was not aligned (i.e. different seeds created different amounts of data). The plot shows the mean (bold lines) and standard deviation (shaded area), averaged across all seeds and data points in the same bin.

We find that across seeds and collect intervals, our method outperforms the vanilla PlaNet implementation at the same time steps. The observed reward showed high variance which was also the case in the original PlaNet work. In simulation, more time for model fitting in the original PlaNet algorithm (i.e. higher collect interval value) was associated with higher performance, but to our surprise, on the real robot, one specific collect interval setting outperformed the others. We conjecture that this is due to the model overfitting to the non-stationary movements of the low-cost robot platform that are present in the real system but not in simulation.

\section{Conclusion \& Future Work}

We introduce Robo-PlaNet, an asynchronous extension of the PlaNet architecture and demonstrate that it speeds up training by minimizing robot downtime. Through this work, we hope to have created a more practical variant of PlaNet.

Our current approach has several limitations. In this work, we have only shown performance on one task and on one robot. In the future, we would like to extend this to more robotic tasks. Like PlaNet, we rely on the environment delivering a dense reward signal for a task, which severely limits the applicability of this method. The original PlaNet paper includes a single task with sparse reward which is solvable in a fraction of the planning horizon. We believe that in order to create a method that allows training a robot from scratch, a better method for learning a reward signal is needed. \citet{singh2019} have recently shown some progress in that direction. 

We also believe that most \ac{RL} methods can be accelerated through the help of human demonstrations. Since a human experimenter has to be present for much of the robot's exploration, learning from demonstration would be a natural fit for this problem. Similarly, to speed up training even further, multiple robots could be used in an architecture similar to GORILA~\cite{nair2015massively}.

\section*{Acknowledgments}

This research was enabled in part by support provided by Calcul Quebec and Compute Canada. We gratefully acknowledge support provided by NVIDIA who donated GPUs through the NVIDIA GPU Grant program. We thank ElementAI for granting us access to their equipment. We would also like to thank INRIA Bordeaux (FLOWERS team) for the development of the robots and the continuing support, Prof. Liam Paull and Prof. David Meger for their advice, and Bhairav Mehta for his ongoing feedback. 

\printbibliography
\end{document}